\documentclass{article}
\usepackage{ijcai21}

\usepackage{times}
\usepackage{soul}
\usepackage{url}
\usepackage[hidelinks]{hyperref}
\usepackage[utf8]{inputenc}
\usepackage[small]{caption}
\usepackage{graphicx}
\usepackage{subfigure}
\usepackage{amsmath}
\usepackage{amsthm}
\usepackage{amsfonts}
\usepackage{booktabs}
\usepackage{multirow}
\usepackage{algorithm}
\usepackage{algorithmic}
\newtheorem{theorem}{Theorem}
\newtheorem{lemma}[theorem]{Lemma}

\title{Cross-Domain Few-Shot Classification via Adversarial Task Augmentation}

\author{
Haoqing Wang\and
Zhi-Hong Deng\footnote{Corresponding author}\\
\affiliations
School of Electronics Engineering and Computer Science, Peking University, Beijing, China\\
\emails
wanghaoqing@pku.edu.cn\and
zhdeng@pku.edu.cn
}

\begin{document}

\maketitle

\begin{abstract}
Few-shot classification aims to recognize unseen classes with few labeled samples from each class. Many meta-learning models for few-shot classification elaborately design various task-shared inductive bias (meta-knowledge) to solve such tasks, and achieve impressive performance. However, when there exists the domain shift between the training tasks and the test tasks, the obtained inductive bias fails to generalize across domains, which degrades the performance of the meta-learning models. In this work, we aim to improve the robustness of the inductive bias through task augmentation. Concretely, we consider the worst-case problem around the source task distribution, and propose the adversarial task augmentation method which can generate the inductive bias-adaptive 'challenging' tasks. Our method can be used as a simple plug-and-play module for various meta-learning models, and improve their cross-domain generalization capability. We conduct extensive experiments under the cross-domain setting, using nine few-shot classification datasets: mini-ImageNet, CUB, Cars, Places, Plantae, CropDiseases, EuroSAT, ISIC and ChestX. Experimental results show that our method can effectively improve the few-shot classification performance of the meta-learning models under domain shift, and outperforms the existing works. Our code is available at \url{https://github.com/Haoqing-Wang/CDFSL-ATA}.
\end{abstract}

\section{Introduction}
Few-shot classification \cite{lake2015human} aims to classify instances from unseen classes with few labeled samples in each class. To this end, many meta-learning based models elaborately design various task-shared inductive bias (e.g., the metric function \cite{sung2018learning}, the inference mechanism \cite{garcia2018few,liu2019learning}) to solve few-shot classification tasks. They demonstrate promising performance when evaluated on the tasks from the same domain with the training tasks (e.g., both training and testing are on the mini-ImageNet classes). However, some works \cite{DBLP:conf/iclr/ChenLKWH19,guo2020broader} have shown that the existing meta-learning models perform undesirably when there exists domain shift between training tasks and test tasks (e.g., training on the mini-ImageNet classes and testing on the ISIC classes), and even underperform compared to traditional pre-training and fine-tuning. As a result, the cross-domain few-shot classification problem has attracted considerable attention from the machine learning community, especially the difficult \emph{single} domain generalization problem \cite{DBLP:conf/iclr/TsengLH020,guo2020broader}.

\begin{figure}
    \centering
    \subfigure[]{
        \includegraphics[width=0.2\textwidth]{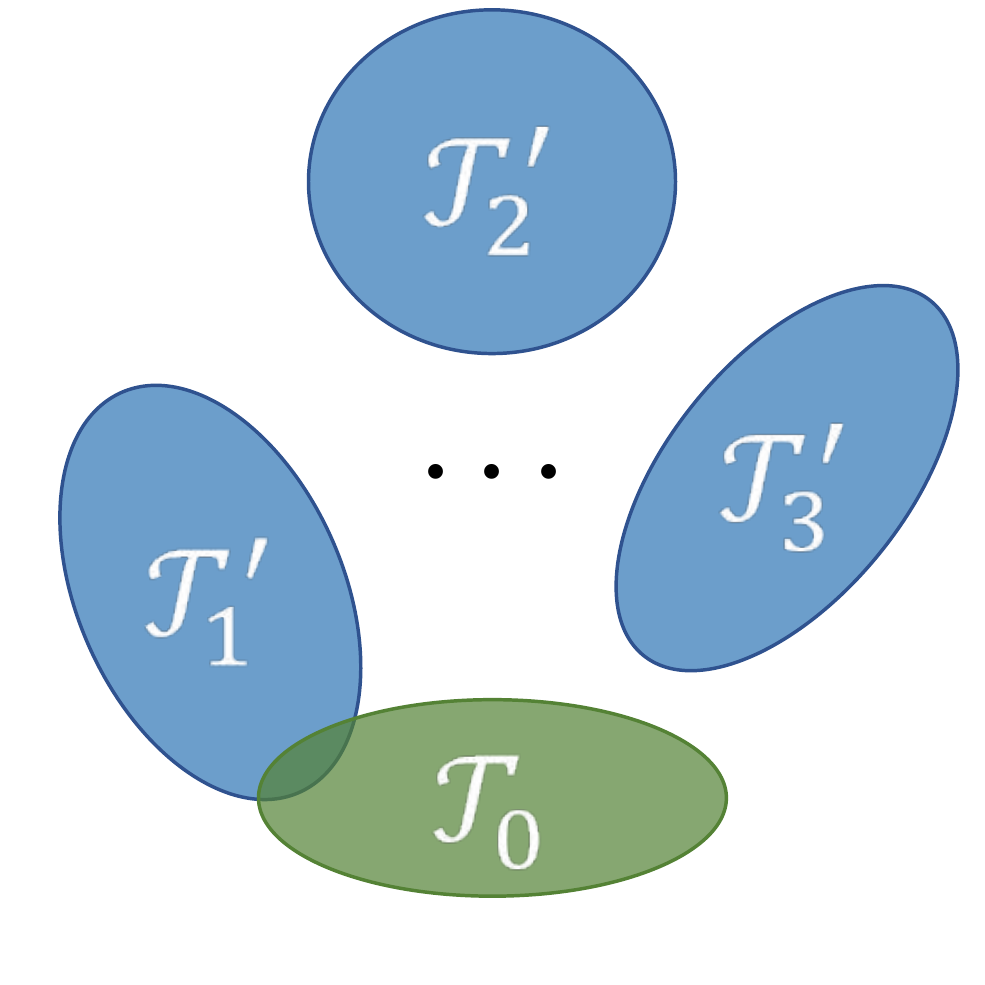}
        \label{fig1a}}
    \subfigure[]{
        \includegraphics[width=0.2\textwidth]{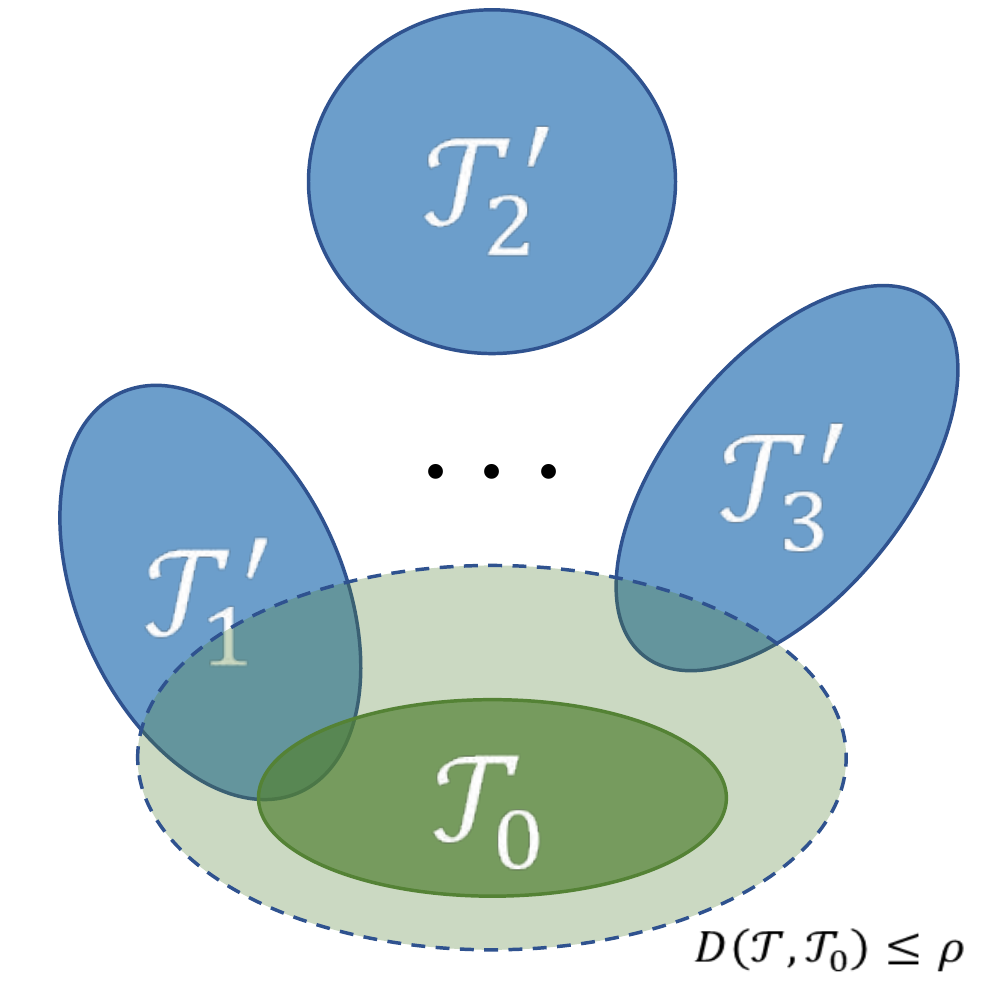}
        \label{fig1b}}
    \caption{Compared with (a) generalizing from the single source task distribution $\mathcal{T}_0$, (b) the worst-case problem considers the wider task distribution space $\{\mathcal{T}|D(\mathcal{T},\mathcal{T}_0)\leq\rho\}$. $\mathcal{T}_1'$, $\mathcal{T}_2'$ and $\mathcal{T}_3'$ represent the unknown task distributions.}
    \label{fig1}
\end{figure}

To generalize to unseen domains without accessing any data from those domains, some domain generalization models have been proposed \cite{volpi2018generalizing,DBLP:conf/icml/LiYZH19}. They learn the classifiers that generalize to the unseen domains, and assume that the source and unseen domains share the same classes. However, in the few-shot classification problem, the classes in the target tasks are unseen before. The most similar works to ours are \cite{DBLP:conf/iclr/TsengLH020} and \cite{sun2020explanation} which aim to improve the performance of meta-learning models in cross-domain tasks. \cite{DBLP:conf/iclr/TsengLH020} introduces the feature-wise transformation layers for the metric-based meta-learning models which modulate the feature activation with affine transformation to improve the robustness of the metric functions. But as mentioned above, the different meta-learning models have various inductive bias, not just the metric functions. \cite{sun2020explanation}
uses explanation-guided training to prevent the feature extractor from overfitting to specific classes, but it needs to manually derive the explanations for different meta-learning models.

We aim to find a method that is general, easy to implement and can improve the robustness of various inductive bias. To this end, we resort to the task augmentation techniques which constructs 'challenging' virtual tasks to increase the diversity of training tasks. For image classification, various hand-crafted data augmentation techniques (e.g., horizontal flip, random crop and color jitter) can be used for task augmentation. However, they have limited effect and cannot perform adaptive augmentation for different inductive bias. Recently, some works \cite{DBLP:conf/iclr/SinhaND18,volpi2018generalizing} proposed adaptive sample (e.g., images) augmentation methods to improve the robustness of the model. Inspired by these works, we propose an inductive bias-adaptive task augmentation method to improve the cross-domain generalization ability of the meta-learning models.

Concretely, we consider the worst-case problem around the source task distribution $\mathcal{T}_0$
\begin{equation}\label{wc1}
    \min_{\theta\in\Theta}\sup_{D(\mathcal{T},\mathcal{T}_0)\leq\rho}\mathbb{E}_{\mathcal{T}}[L(T;\theta)]
\end{equation}
where $\theta\in\Theta$ represents the model parameters, $L(T;\theta)$ is the loss function which depends on the model's inductive bias, and $D(\mathcal{T},\mathcal{T}_0)$ is the distance metric between task distributions. Compared with minimizing the loss function on the source task distribution $\mathcal{T}_0$, the solution to the worst-case problem (\ref{wc1}) guarantees good performance on the wider space of task distributions which are $\rho$ distance away from $\mathcal{T}_0$, as illustrated in Figure \ref{fig1}. By solving the worst-case problem (\ref{wc1}), we propose a task augmentation method. Since the loss function depends on the inductive bias, our method can adaptively generate 'challenging' tasks according to the different inductive bias and increase the diversity of training tasks which improves the robustness of the model under domain shift. What's more, our method can be used as a plug-and-play module for various meta-learning models.

The main contributions of this work are as follows:
\begin{itemize}
\item To the best of our knowledge, this is the first work that introduces task augmentation into cross-domain few-shot classification to improve the generalization ability of meta-learning models under domain shift.
\item We consider the worst-case problem around the source task distribution $\mathcal{T}_0$, and propose a plug-and-play inductive bias-adaptive task augmentation method, which can be conveniently used for various meta-learning models.
\item We evaluate our method on the RelationNet \cite{sung2018learning}, the GNN \cite{garcia2018few} and one of the state-of-the-art models TPN \cite{liu2019learning} with extensive experiments under the cross-domain setting. Experimental results show our method can significantly improve the cross-domain generalization performance of these models and outperforms \cite{DBLP:conf/iclr/TsengLH020} and \cite{sun2020explanation}. And under the same settings, the meta-learning models with our adversarial task augmentation module can outperform the traditional pre-training and fine-tuning under domain shift.
\end{itemize}

\section{Related Work}
\paragraph{Cross-domain few-shot classification.} Although various meta-learning models for few-shot classification have achieved impressive performance, they fail to generalize to unseen domains. To this end, \cite{DBLP:conf/iclr/TsengLH020} uses the feature-wise transformation layers to simulate various distributions of image features during training and thus improve the generalization capability of the metric function. \cite{sun2020explanation} uses the explanation methods to upscale the features which are more relevant to the prediction, and penalize them more when overfitting occurs to avoid the intermediate features from specializing towards fixed classes. Different from them, we focus on improving the robustness of various inductive bias. Other models \cite{liu2020feature,yeh2020large} that appear in the CVPR 2020 Cross-Domain Few-Shot Learning Challenge use various techniques to solve cross-domain few-shot classification tasks, e.g., batch spectral regularization, model ensemble and large margin mechanism.

\paragraph{Domain generalization.} Domain generalization methods \cite{volpi2018generalizing,DBLP:conf/icml/LiYZH19} have been developed to generalizing from single or multiple seen domains to the unseen domains without accessing samples from them. However, these models consider the setting that the seen and unseen domains share the same categories. In contrast, in the cross-domain few-shot classification problem, the seen and the unseen domains have completely disjoint categories.

\paragraph{Adversarial training.} Adversarial training \cite{DBLP:journals/corr/GoodfellowSS14} aims to make deep neural networks be capable of resistant to adversarial attacks. \cite{DBLP:conf/iclr/SinhaND18} proposes principled adversarial training through distributionally robust optimization, where virtual images are model-adaptively generated by maximize some risk and the models learned with these new images become more robust. In this work, we introduce a similar model-adaptive augmentation method into the meta-learning models, and propose a plug-and-play module to generate virtual 'challenging' tasks to improve the robustness of various meta-learning models.

\section{Method}
\subsection{Preliminaries}
\subsubsection{Few-Shot Classification} 
Each few-shot classification task $T$ consists of a support set $T_s$ and a query set $T_q$. If the support set $T_s$ contains $C$ classes with $K$ samples in each class, the few-shot classification task is called $C$-way $K$-shot. The query set $T_q$ contains the samples from the same classes with the support set $T_s$. Formally, a few-shot task can be defined as $T=(T_s,T_q)$, where $T_s=\{(x_i^s,y_i^s)\}_{i=1}^{C\times K}$ and $T_q=\{(x_j^q,y_j^q)\}_{j=1}^{Q}$. Given the support set $T_s$, our goal is to classify the samples in the query set $T_q$ correctly to one of the $C$ classes. Typically, the base learner $\mathcal{A}$ is needed to output the optimal classifier $\psi$ of the task basing on the support set $T_s$, i.e., $\psi=\mathcal{A}(T_s;\theta)$ and it depends on the inductive bias.

The main difference among meta-learning models for few-shot classification lies in the design choices for the inductive bias. For examples, the RelationNet \cite{sung2018learning} chooses the metric function based on convolutional neural networks (CNNs), the GNN \cite{garcia2018few} applies generic message-passing inference mechanism on a partially observed graphical model, and the TPN \cite{liu2019learning} utilizes the transductive label propagation. Meta-learning models aim to learn these inductive bias over a collection of tasks which are assumed to be sampled from the task distribution $\mathcal{T}_0$, and the learning objective is
\begin{equation}\label{ml1}
    \min_{\theta\in\Theta}\mathbb{E}_{(T_s,T_q)\sim\mathcal{T}_0}[L^{meta}(T_q,\psi)],\psi=\mathcal{A}(T_s;\theta)
\end{equation}
where $L^{meta}$ is the loss function, such as the classification loss of the samples in the query set $T_q$, and $\theta$ represents the model parameters.

\subsubsection{Cross-Domain Setting}
Generally, the target tasks are assumed to come from the source task distribution $\mathcal{T}_0$. However, in this work we consider the few-shot classification under domain shift. Concretely, we focus on the \emph{single} domain generalization problem because the data from multiple training domains may not always be available due to data acquiring budget or privacy issue. We denote the domain as the distribution of the few-shot classification tasks. The target tasks come from several unknown domains $\{\mathcal{T}_1',\cdots,\mathcal{T}_N'\}$. The goal is to learn a meta-learning model using the single source domain $\mathcal{T}_0$, such that the model can generalize to the several unseen domains.

\begin{algorithm}[tb]
\caption{Adversarial Task Augmentation}
\label{ATA}
\textbf{Input}: Source task distribution $\mathcal{T}_0$; initialized parameters $\theta_0$ \\
\textbf{Require}: Learning rate $\alpha$ and $\beta$; iteration number for early stopping $\mathbf{T}_{max}$; probability of using original data $p\in(0,1)$; candidate pool of filter sizes $\mathcal{K}$ \\
\textbf{Output}: learned parameters $\theta$
\begin{algorithmic}[1]
\STATE \textbf{Initialize}: $\theta\leftarrow\theta_0$
\WHILE{training}
\STATE Randomly sample source task $T_0=(X_0,Y_0)$ from $\mathcal{T}_0$
\STATE $X_0\leftarrow RandConv(X_0, \mathcal{K})$ ( with probability $1-p$ )
\FOR{$i=1,\dots,\mathbf{T}_{max}$}
\STATE $X_i=X_{i-1}+\beta\cdot\nabla_XL((X_{i-1},Y_0);\theta)$
\ENDFOR
\STATE $\theta\leftarrow\theta-\alpha\nabla_{\theta}L((X_{\mathbf{T}_{max}},Y_0);\theta)$
\ENDWHILE
\end{algorithmic}
\end{algorithm}

\begin{table*}[ht]
\centering
\scalebox{0.96}{
\begin{tabular}{ccccccccc}
\toprule
\multicolumn{1}{c}{\multirow{2}{*}{Model}}&\multicolumn{2}{c}{CUB}&\multicolumn{2}{c}{Cars}&\multicolumn{2}{c}{Places}&\multicolumn{2}{c}{Plantae}
\\ \cmidrule(lr){2-3} \cmidrule(lr){4-5} \cmidrule(lr){6-7} \cmidrule(lr){8-9}
\multicolumn{1}{c}{} & \multicolumn{1}{c}{1-shot}&\multicolumn{1}{c}{5-shot} & \multicolumn{1}{c}{1-shot}&\multicolumn{1}{c}{5-shot} & \multicolumn{1}{c}{1-shot}&\multicolumn{1}{c}{5-shot} & \multicolumn{1}{c}{1-shot}&\multicolumn{1}{c}{5-shot} \\
\hline
RelationNet     & $41.27_{\pm0.4}$ & $56.77_{\pm0.4}$ & $30.09_{\pm0.3}$ & $40.46_{\pm0.4}$ & $48.16_{\pm0.5}$ & $64.25_{\pm0.4}$ & $31.23_{\pm0.3}$ & $42.71_{\pm0.3}$ \\
+FT \cite{DBLP:conf/iclr/TsengLH020} & $\textbf{43.33}_{\pm\textbf{0.4}}$ & $\textbf{59.77}_{\pm\textbf{0.4}}$ & $30.45_{\pm0.3}$ & $40.18_{\pm0.4}$ & $49.92_{\pm0.5}$ & $65.55_{\pm0.4}$ & $32.57_{\pm0.3}$ & $44.29_{\pm0.3}$ \\
+LRP \cite{sun2020explanation} & $41.57_{\pm0.4}$ & $57.70_{\pm0.4}$ & $30.48_{\pm0.3}$ & $41.21_{\pm0.4}$ & $48.47_{\pm0.5}$ & $65.35_{\pm0.4}$ & $32.11_{\pm0.3}$ & $43.70_{\pm0.3}$ \\
+ATA (Ours) & $\textbf{43.02}_{\pm\textbf{0.4}}$ & $\textbf{59.36}_{\pm\textbf{0.4}}$ & $\textbf{31.79}_{\pm\textbf{0.3}}$ & $\textbf{42.95}_{\pm\textbf{0.4}}$ & $\textbf{51.16}_{\pm\textbf{0.5}}$ & $\textbf{66.90}_{\pm\textbf{0.4}}$ & $\textbf{33.72}_{\pm\textbf{0.3}}$ & $\textbf{45.32}_{\pm\textbf{0.3}}$ \\
\hline
GNN             & $44.40_{\pm0.5}$ & $62.87_{\pm0.5}$ & $31.72_{\pm0.4}$ & $43.70_{\pm0.4}$ & $52.42_{\pm0.5}$ & $70.91_{\pm0.5}$ & $33.60_{\pm0.4}$ & $48.51_{\pm0.4}$ \\
+FT \cite{DBLP:conf/iclr/TsengLH020} & $\textbf{45.50}_{\pm\textbf{0.5}}$ & $64.97_{\pm0.5}$ & $32.25_{\pm0.4}$ & $46.19_{\pm0.4}$ & $\textbf{53.44}_{\pm\textbf{0.5}}$ & $70.70_{\pm0.5}$ & $32.56_{\pm0.4}$ & $49.66_{\pm0.4}$ \\
+LRP \cite{sun2020explanation} & $43.89_{\pm0.5}$ & $62.86_{\pm0.5}$ & $31.46_{\pm0.4}$ & $46.07_{\pm0.4}$ & $52.28_{\pm0.5}$ & $71.38_{\pm0.5}$ & $33.20_{\pm0.4}$ & $50.31_{\pm0.4}$ \\
+ATA (Ours) & $\textbf{45.00}_{\pm\textbf{0.5}}$ & $\textbf{66.22}_{\pm\textbf{0.5}}$ & $\textbf{33.61}_{\pm\textbf{0.4}}$ & $\textbf{49.14}_{\pm\textbf{0.4}}$ & $\textbf{53.57}_{\pm\textbf{0.5}}$ & $\textbf{75.48}_{\pm\textbf{0.4}}$ & $\textbf{34.42}_{\pm\textbf{0.4}}$ & $\textbf{52.69}_{\pm\textbf{0.4}}$ \\
\hline
TPN             & $48.03_{\pm0.4}$ & $63.52_{\pm0.4}$ & $32.42_{\pm0.4}$ & $44.54_{\pm0.4}$ & $\textbf{56.17}_{\pm\textbf{0.5}}$ & $\textbf{71.39}_{\pm\textbf{0.4}}$ & $37.40_{\pm0.4}$ & $50.96_{\pm0.4}$ \\
+FT \cite{DBLP:conf/iclr/TsengLH020} & $44.24_{\pm0.5}$ & $58.18_{\pm0.5}$ & $26.50_{\pm0.3}$ & $34.03_{\pm0.4}$ & $52.45_{\pm0.5}$ & $66.75_{\pm0.5}$ & $32.46_{\pm0.4}$ & $43.20_{\pm0.5}$ \\
+ATA (Ours) & $\textbf{50.26}_{\pm\textbf{0.5}}$ & $\textbf{65.31}_{\pm\textbf{0.4}}$ & $\textbf{34.18}_{\pm\textbf{0.4}}$ & $\textbf{46.95}_{\pm\textbf{0.4}}$ & $\textbf{57.03}_{\pm\textbf{0.5}}$ & $\textbf{72.12}_{\pm\textbf{0.4}}$ & $\textbf{39.83}_{\pm\textbf{0.4}}$ & $\textbf{55.08}_{\pm\textbf{0.4}}$ \\
\midrule
\multicolumn{1}{c}{\multirow{2}{*}{}}&\multicolumn{2}{c}{CropDiseases}&\multicolumn{2}{c}{EuroSAT}&\multicolumn{2}{c}{ISIC}&\multicolumn{2}{c}{ChestX}
\\ \cmidrule(lr){2-3} \cmidrule(lr){4-5} \cmidrule(lr){6-7} \cmidrule(lr){8-9}
\multicolumn{1}{c}{} & \multicolumn{1}{c}{1-shot}&\multicolumn{1}{c}{5-shot} & \multicolumn{1}{c}{1-shot}&\multicolumn{1}{c}{5-shot} & \multicolumn{1}{c}{1-shot}&\multicolumn{1}{c}{5-shot} & \multicolumn{1}{c}{1-shot}&\multicolumn{1}{c}{5-shot} \\
\hline
RelationNet     & $53.58_{\pm0.4}$ & $72.86_{\pm0.4}$ & $49.08_{\pm0.4}$ & $65.56_{\pm0.4}$ & $30.53_{\pm0.3}$ & $38.60_{\pm0.3}$ & $21.95_{\pm0.2}$ & $\textbf{24.07}_{\pm\textbf{0.2}}$ \\
+FT \cite{DBLP:conf/iclr/TsengLH020}  & $57.57_{\pm0.5}$ & $75.78_{\pm0.4}$ & $53.53_{\pm0.4}$ & $69.13_{\pm0.4}$ & $30.38_{\pm0.3}$ & $38.68_{\pm0.3}$ & $\textbf{21.79}_{\pm\textbf{0.2}}$ & $23.95_{\pm0.2}$ \\
+LRP \cite{sun2020explanation} & $55.01_{\pm0.4}$ & $74.21_{\pm0.4}$ & $50.99_{\pm0.4}$ & $67.54_{\pm0.4}$ & $\textbf{31.16}_{\pm\textbf{0.3}}$ & $\textbf{39.97}_{\pm\textbf{0.3}}$ & $\textbf{22.11}_{\pm\textbf{0.2}}$ & $\textbf{24.28}_{\pm\textbf{0.2}}$ \\
+ATA (Ours) & $\textbf{61.17}_{\pm\textbf{0.5}}$ & $\textbf{78.20}_{\pm\textbf{0.4}}$ & $\textbf{55.69}_{\pm\textbf{0.5}}$ & $\textbf{71.02}_{\pm0.4}$ & $\textbf{31.13}_{\pm\textbf{0.3}}$ & $\textbf{40.38}_{\pm\textbf{0.3}}$ & $\textbf{22.14}_{\pm\textbf{0.2}}$ & $\textbf{24.43}_{\pm\textbf{0.2}}$ \\
\hline
GNN             & $59.19_{\pm0.5}$ & $83.12_{\pm0.4}$ & $54.61_{\pm0.5}$ & $78.69_{\pm0.4}$ & $30.14_{\pm0.3}$ & $42.54_{\pm0.4}$ & $\textbf{21.94}_{\pm\textbf{0.2}}$ & $23.87_{\pm0.2}$ \\
+FT \cite{DBLP:conf/iclr/TsengLH020} & $60.74_{\pm0.5}$ & $87.07_{\pm0.4}$ & $55.53_{\pm0.5}$ & $78.02_{\pm0.4}$ & $30.22_{\pm0.3}$ & $40.87_{\pm0.4}$ & $\textbf{22.00}_{\pm\textbf{0.2}}$ & $\textbf{24.28}_{\pm\textbf{0.2}}$ \\
+LRP \cite{sun2020explanation} & $59.23_{\pm0.5}$ & $86.15_{\pm0.4}$ & $54.99_{\pm0.5}$ & $77.14_{\pm0.4}$ & $30.94_{\pm0.3}$ & $\textbf{44.14}_{\pm\textbf{0.4}}$ & $\textbf{22.11}_{\pm\textbf{0.2}}$ & $\textbf{24.53}_{\pm\textbf{0.3}}$ \\
+ATA (Ours) & $\textbf{67.47}_{\pm\textbf{0.5}}$ & $\textbf{90.59}_{\pm\textbf{0.3}}$ & $\textbf{61.35}_{\pm\textbf{0.5}}$ & $\textbf{83.75}_{\pm\textbf{0.4}}$ & $\textbf{33.21}_{\pm\textbf{0.4}}$ & $\textbf{44.91}_{\pm\textbf{0.4}}$ & $\textbf{22.10}_{\pm\textbf{0.2}}$ & $\textbf{24.32}_{\pm\textbf{0.4}}$ \\
\hline
TPN             & $68.39_{\pm0.6}$ & $81.91_{\pm0.5}$ & $63.90_{\pm0.5}$ & $77.22_{\pm0.4}$ & $\textbf{35.08}_{\pm\textbf{0.4}}$ & $\textbf{45.66}_{\pm\textbf{0.3}}$ & $21.05_{\pm0.2}$ & $22.17_{\pm0.2}$ \\
+FT \cite{DBLP:conf/iclr/TsengLH020} & $56.06_{\pm0.7}$ & $70.06_{\pm0.7}$ & $52.68_{\pm0.6}$ & $65.69_{\pm0.5}$ & $29.62_{\pm0.3}$ & $36.96_{\pm0.4}$ & $20.46_{\pm0.1}$ & $21.22_{\pm0.1}$ \\
+ATA (Ours) & $\textbf{77.82}_{\pm\textbf{0.5}}$ & $\textbf{88.15}_{\pm\textbf{0.5}}$ & $\textbf{65.94}_{\pm\textbf{0.5}}$ & $\textbf{79.47}_{\pm\textbf{0.3}}$ & $\textbf{34.70}_{\pm\textbf{0.4}}$ & $\textbf{45.83}_{\pm\textbf{0.3}}$ & $\textbf{21.67}_{\pm\textbf{0.2}}$ & $\textbf{23.60}_{\pm\textbf{0.2}}$ \\
\bottomrule
\end{tabular}}
\caption{Few-shot classification accuracy$(\%)$ of 5-way 1-shot/5-shot tasks trained with the mini-ImageNet dataset. \textbf{+FT} means using the feature-wise transformation layers, \textbf{+LRP} means using the explanation-guided training, \textbf{+ATA} means using our adversarial task augmentation. Marked in bold are the best results in each block, as well as other results with an overlapping confidence interval.}
\label{SOTA}
\end{table*}

\subsection{Adversarial Task Augmentation}\label{sec32}
Next, we solve the worst-case problem (\ref{wc1}) to get a plug-and-play model-adaptive task augmentation module. In order to make the loss function $L(T;\theta)$ depending on the inductive bias of the meta-learning models, inspired by Equation (\ref{ml1}), we define it as
\begin{equation}\label{loss}
  L(T;\theta)=L^{meta}(T_q,\psi),\psi=\mathcal{A}(T_s;\theta)
\end{equation}
To allow task distributions that have different support to that of the source task distribution $\mathcal{T}_0$, we use the Wasserstein distances as the metric $D$. Concretely, for task distribution $\mathcal{T}$ and $\mathcal{T}_0$ both supported on the task space $\mathcal{H}$, let $\Pi(\mathcal{T},\mathcal{T}_0)$ denotes their couplings, meaning measures $M$ on $\mathcal{H}^2$ with $M(T,\mathcal{H})=\mathcal{T}(T)$ and $M(\mathcal{H},T)=\mathcal{T}_0(T)$. The Wasserstein distance between $\mathcal{T}$ and $\mathcal{T}_0$ is
\begin{equation}\label{dis}
    D(\mathcal{T},\mathcal{T}_0)=\inf_{M\in\Pi(\mathcal{T},\mathcal{T}_0)}\mathbb{E}_M[d(T,T_0)]
\end{equation}
where $d:\mathcal{H}\times\mathcal{H}\rightarrow\mathbb{R}_+$ is the transportation cost from $T$ to $T_0$, satisfying $d(T,T_0)\geq0$ and $d(T,T)=0$.

Basing on the Proposition 1 in \cite{DBLP:conf/iclr/SinhaND18} and the Theorem 1 in \cite{blanchet2019quantifying}, we have the following duality result.
\begin{lemma}\label{lm1}
Let $L:\mathcal{H}\times\Theta\rightarrow\mathbb{R}$ and $d:\mathcal{H}\times\mathcal{H}\rightarrow\mathbb{R}_+$ be continuous. Let $\phi_{\gamma}(T_0;\theta)=\sup_{T\in\mathcal{H}}\{L(T;\theta)-\gamma d(T,T_0)\}$ be the cross domain surrogate. For any distribution $\mathcal{T}_0$ and any $\rho>0$,
\begin{equation}
    \sup_{D(\mathcal{T},\mathcal{T}_0)\leq\rho}\mathbb{E}_{\mathcal{T}}[L(T;\theta)]=\inf_{\gamma\geq0}\{\gamma\rho+\mathbb{E}_{\mathcal{T}_0}[\phi_{\gamma}(T_0;\theta)]\}
\end{equation}
and for any $\gamma\geq0$, we have
\begin{equation}
    \sup_{\mathcal{T}}\{\mathbb{E}_{\mathcal{T}}[L(T;\theta)]-\gamma D(\mathcal{T},\mathcal{T}_0)\}=\mathbb{E}_{\mathcal{T}_0}[\phi_{\gamma}(T_0;\theta)]
\end{equation}
\end{lemma}

Thus, the continuity of the loss function $L(T;\theta)$ and the transportation function $d(T;T_0)$ with respect to $T$ needs to be satisfied to solve the worst-case problem (\ref{wc1}). For this, we model the task $T$ as the vector with the fixed dimension. A common approach is to use task embedding to model the tasks, but it is not applicable here. The reasons are as follows: 1) it conflicts with the definition of the loss function $L(T;\theta)$, i.e., calculating $L(T;\theta)$ requires the support set $T_s$ and query set $T_q$, not the task embedding; 2) we expect $\mathcal{T}_0$ and $\mathcal{T}$ to be the distribution of the tasks to generate virtual tasks not the task embedding. We treat each task as the vector concatenated by the samples and labels it contains, i.e.,
\begin{equation}\label{task}
    T=[x_1^s,y_1^s,\cdots,x_{C\times K}^s,y_{C\times K}^s,x_1^q,y_1^q,\cdots,x_{Q}^q,y_{Q}^q]
\end{equation}
where $[\cdot,\cdot]$ denotes the concatenation operation. This definition is equivalent to treating the distribution of the tasks as the joint distribution of samples and labels within the task, i.e.
\begin{equation}
\resizebox{.91\linewidth}{!}{$\mathcal{T}(T)=P(x_1^s,y_1^s,\cdots,x_{C\times K}^s,y_{C\times K}^s,x_1^q,y_1^q,\cdots,x_{Q}^q,y_{Q}^q)$}
\end{equation}
Meanwhile, we assume that the number of samples in the task $T$ is fixed, so as the dimension of $T$. The change of the elements of the samples in task $T$ leads to the change of $T$, so the continuity of $L(T;\theta)$ and $d(T;T_0)$ can be satisfied. Another consideration for assuming a fixed number of samples in a task is that we want to generate the virtual task containing the same number of samples with source task $T$.

In the worst-case problem (\ref{wc1}), the supremum over task distributions is intractable, so we consider its Lagrangian relaxation with penalty parameter $\gamma\geq0$
\begin{equation}
   \min_{\theta\in\Theta} \sup_{\mathcal{T}}\{\mathbb{E}_{\mathcal{T}}[L(T;\theta)]-\gamma D(\mathcal{T},\mathcal{T}_0)\}
\end{equation}
Applying Lemma \ref{lm1}, our optimization problem becomes
\begin{equation}\label{wc2}
   \min_{\theta\in\Theta}\mathbb{E}_{\mathcal{T}_0}[\phi_{\gamma}(T_0;\theta)]
\end{equation}
Further, applying the Theorem 4.13 in \cite{bonnans2013perturbation}, we have the following result to solve the problem (\ref{wc2}).
\begin{lemma}\label{lm2}
Let $L:\mathcal{H}\times\Theta\rightarrow\mathbb{R}$ be $\lambda$-Lipschitz smooth and $d(\cdot,T_0)$ be $\mu$-strongly convex for each $T_0\in\mathcal{H}$. If $\gamma>\frac{\lambda}{\mu}$, there is unique $\hat{T}$ satisfying
\begin{equation}\label{xn}
    \hat{T}=\arg\sup_{T\in\mathcal{H}}\{L(T;\theta)-\gamma d(T,T_0)\}
\end{equation}
and 
\begin{equation}
    \nabla_{\theta}\phi_{\gamma}(T_0;\theta)=\nabla_{\theta}L(\hat{T};\theta)
\end{equation}
\end{lemma}
In the Lemma \ref{lm2}, $\gamma>\frac{\lambda}{\mu}$ ensures that the function $L(T;\theta)-\gamma d(T,T_0)$ is $(\gamma\mu-\lambda)$-strongly concave in $T$, so that there exists the unique $\hat{T}$.

According to Lemma \ref{lm2}, we can solve the Equation (\ref{xn}) to generate the virtual cross-domain task $\hat{T}$ and use it to update the model parameters $\theta$
\begin{equation}
    \theta \leftarrow \theta - \alpha \nabla_{\theta} L(\hat{T};\theta)
\end{equation}
where $\alpha$ is the learning rate. From the Equation (\ref{xn}), we can make two insights: 1) for the meta-learning models, the virtual task $\hat{T}$ is more 'challenging' than the source task $T_0$ and the loss function satisfies $L(\hat{T};\theta)\geq L(T_0;\theta)+\gamma d(\hat{T},T_0)$, so the model learned with it tends to be more robust; 2) since the loss function $L(T;\theta)$ depends on the inductive bias, solving the Equation (\ref{xn}) is equivalent to adaptively generating the virtual task that is more 'challenging' to the currently learned inductive bias.

For deep networks and other complex models, the supremum problem in Equation (\ref{xn}) cannot be solved accurately, so we use the gradient ascent process with early stopping to solve it. Concretely, let the set of all samples in a task be $X$ and their corresponding labels be $Y$, i.e.,
\begin{align}
  X= & [x_1^s,\cdots,x_{C\times K}^s,x_1^q,\cdots,x_{Q}^q] \\
  Y= & [y_1^s,\cdots,y_{C\times K}^s,y_1^q,\cdots,y_{Q}^q]
\end{align}
then $T=(X,Y)$ and $T_0=(X_0,Y_0)$. We use the source task $T_0$ as the initialization of $T$, and the task vector defined in Equation (\ref{task}) as the optimization variable. Considering that in different few-shot classification tasks, samples with the same labels can correspond to different real category (e.g., cat, dog), so the change of label $Y$ is not considered here, i.e., keeping $Y=Y_0$. In the $i$-th iteration, the update is
\begin{equation}
    X_i=X_{i-1}+\beta\cdot\nabla_XL((X_{i-1},Y_0);\theta)
\end{equation}
Here the regularization term $-\gamma d(T,T_0)$ is removed from the iteration goal and the reasons are as follows: 1) this term is used to constrain the proximity of the virtual task to the source task $T_0$, but using the source task as the initialization and early stopping can achieve the same effect, see the Section \ref{AS} for the detailed discussion; 2) it reduces the computational overhead and hyper-parameters requiring hand-tuning. After $\mathbf{T}_{max}$ iterations, we get the virtual 'challenging' task $\hat{T}=(X_{\mathbf{T}_{max}},Y_0)$ and update the model parameters $\theta$ with it. See Algorithm \ref{ATA} for the full description of the training process. Given an unseen task, the inference process is the same as the original meta-learning model. Note that if $\mathbf{T}_{max}=0$, Algorithm \ref{ATA} becomes the original meta-learning training process, so our method is a plug-and-play module.

In this paper, we mainly consider the cross-domain few-shot \textbf{image} classification, and the convolutional neural networks (CNNs) are the necessary tools. However, CNNs tend to overfit on superficial local textures \cite{DBLP:conf/iclr/GeirhosRMBWB19}, so we use the random convolutions \cite{DBLP:conf/iclr/LeeLSL20} that can change the local textures and keep the shape unchanged as the auxiliary augmentation technique for our adversarial task augmentation. Concretely, given an input image $I\in\mathbb{R}^{C\times H\times W}$, where $H$ and $W$ are the height and width and $C$ is the number of feature channels, the filter size $k$ is first randomly sampled from the candidate pool $\mathcal{K}$, then the Xavier normal distribution \cite{glorot2010understanding} is used to initialize the convolution weights. The stride and padding size are determined to make the transformed image having the same size with $I$. In practice, for each task $T_0$ sampled from $\mathcal{T}_0$, we keep its all samples unchanged with probability $p$, or use the same random convolution on its all samples to get a new task for training, as shown in the fourth line of Algorithm \ref{ATA}.

\begin{figure}
    \centering
    \subfigure[5-way 1-shot setting]{
        \includegraphics[width=0.225\textwidth]{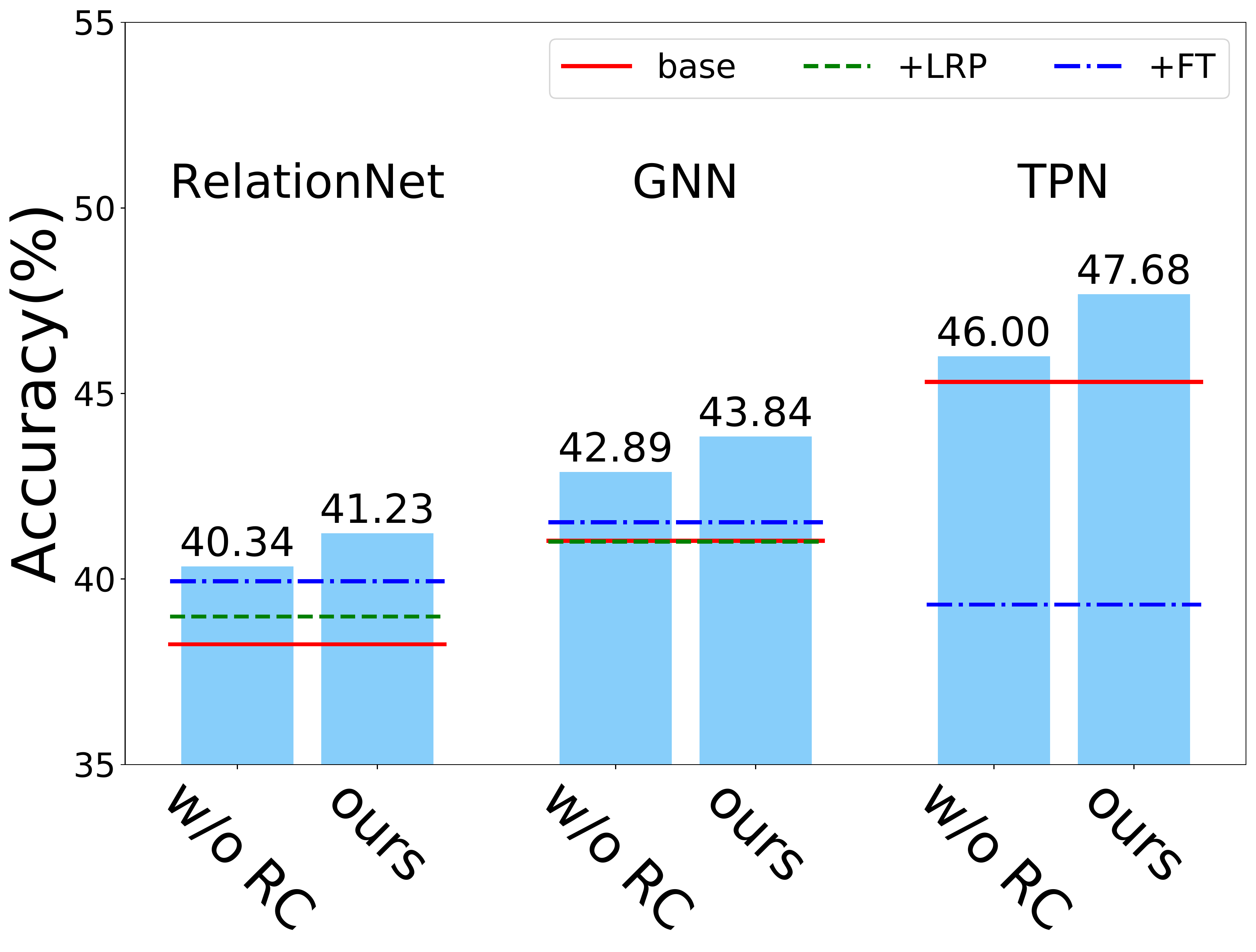}
        \label{fig2a}}
    \subfigure[5-way 5-shot setting]{
        \includegraphics[width=0.225\textwidth]{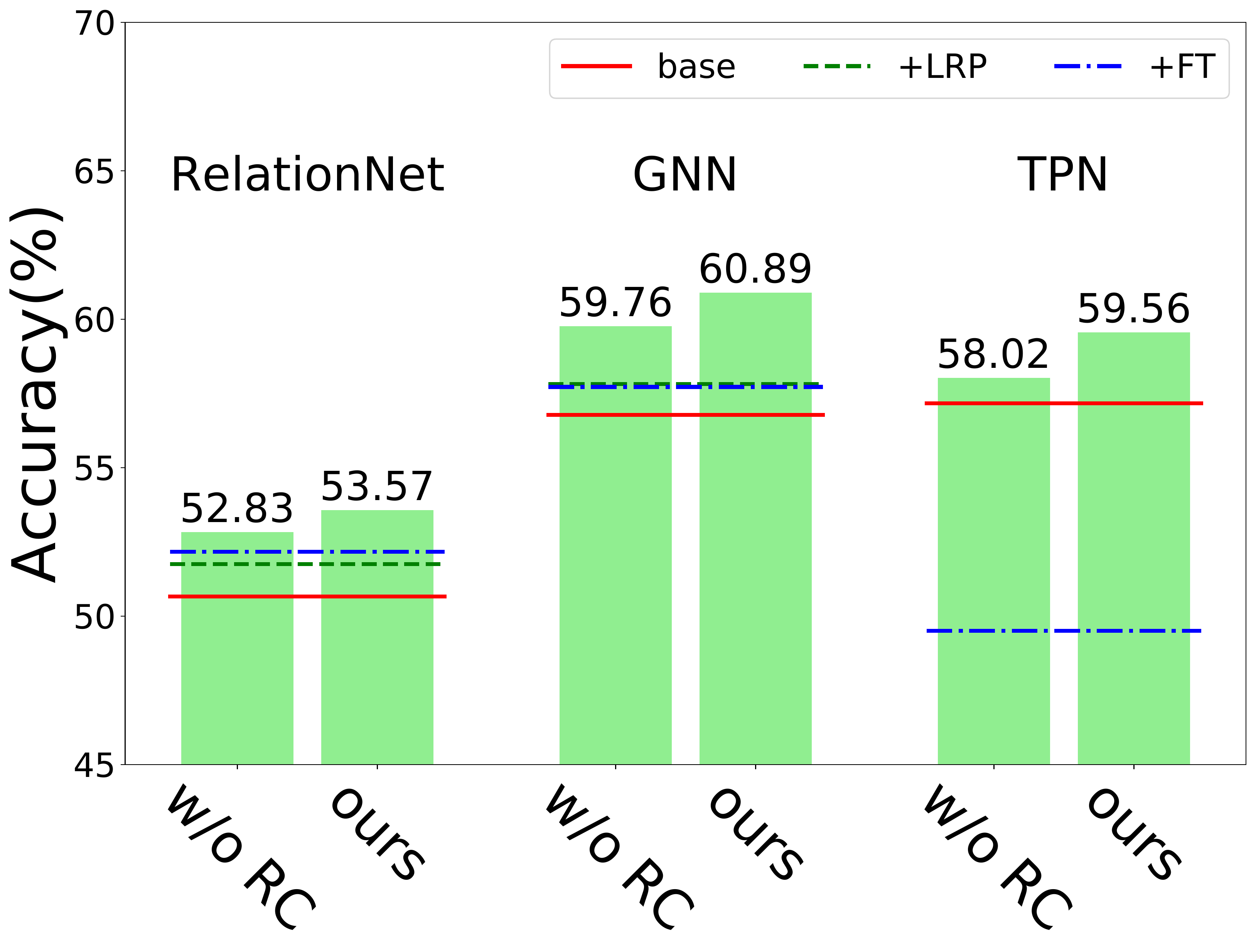}
        \label{fig2b}}
    \caption{Average classification accuracy on eight unseen domains (CUB, Cars, Places, Plantae, CropDiseases, EuroSAT, ISIC and ChestX) under 5-way 1-shot/5-shot setting. It respectively shows the results without the random convolution ('w/o RC') and that obtained by our complete task augmentation module. The results of the base meta-learning models ('base'), the models with the explanation-guided training ('+LRP') and the models with the feature-wise transformation layers ('+FT') are also shown for a clearer comparison.}
    \label{fig2}
\end{figure}

\section{Experiments}
In this section, we evaluate the adversarial task augmentation method on the RelationNet \cite{sung2018learning}, the GNN \cite{garcia2018few} and one of the state-of-the-art meta-learning models TPN \cite{liu2019learning}, and compare it with \cite{DBLP:conf/iclr/TsengLH020} and \cite{sun2020explanation}. These meta-learning models have different kinds of inductive bias so as to verify the versatility and effectiveness of our method.

\subsection{Experimental Settings}
\paragraph{Datasets.} We conduct extensive experiments under cross-domain settings, using nine few-shot classification datasets: mini-ImageNet \cite{DBLP:conf/iclr/RaviL17}, CUB, Cars, Places, Plantae, CropDiseases, EuroSAT, ISIC and ChestX, which are introduced by \cite{DBLP:conf/iclr/TsengLH020} and \cite{guo2020broader}. Each dataset consists of train/val/test splits and please refer to these references for more details. We use the mini-ImageNet domain as the single source domain, and evaluate the trained model on the other eight domains. We select the model parameters with the best accuracy on the validation set of the mini-ImageNet for model evaluation.

\paragraph{Implementation details.} In all experiments, we use the ResNet-10 \cite{he2016deep} as the feature extractor and use the Adam optimizer with the learning rate $\alpha=0.001$. We find that setting $\mathbf{T}_{max}=5$ or $10$ is sufficient to obtain satisfactory results, and we choose the learning rate of the gradient ascent process $\beta$ from $\{20,40,60,80\}$. We set $\mathcal{K}=\{1,3,5,7,11,15\}$ for all experiments and choose $p$ from $\{0.5,0.6,0.7\}$. We evaluate the model in the 5-way 1-shot/5-shot settings using 2,000 randomly sampled episodes with 16 query samples per class, and report the average accuracy ($\%$) as well as $95\%$ confidence interval.

\paragraph{Pre-trained feature extractor.} Instead of optimizing from scratch, we apply an additional pre-training strategy as in \cite{DBLP:conf/iclr/TsengLH020} which pre-trains the feature extractor by minimizing the standard cross-entropy classification loss on the 64 training classes in the mini-ImageNet dataset.

\begin{figure}
    \centering
    \subfigure[5-way 1-shot setting]{
        \includegraphics[width=0.225\textwidth]{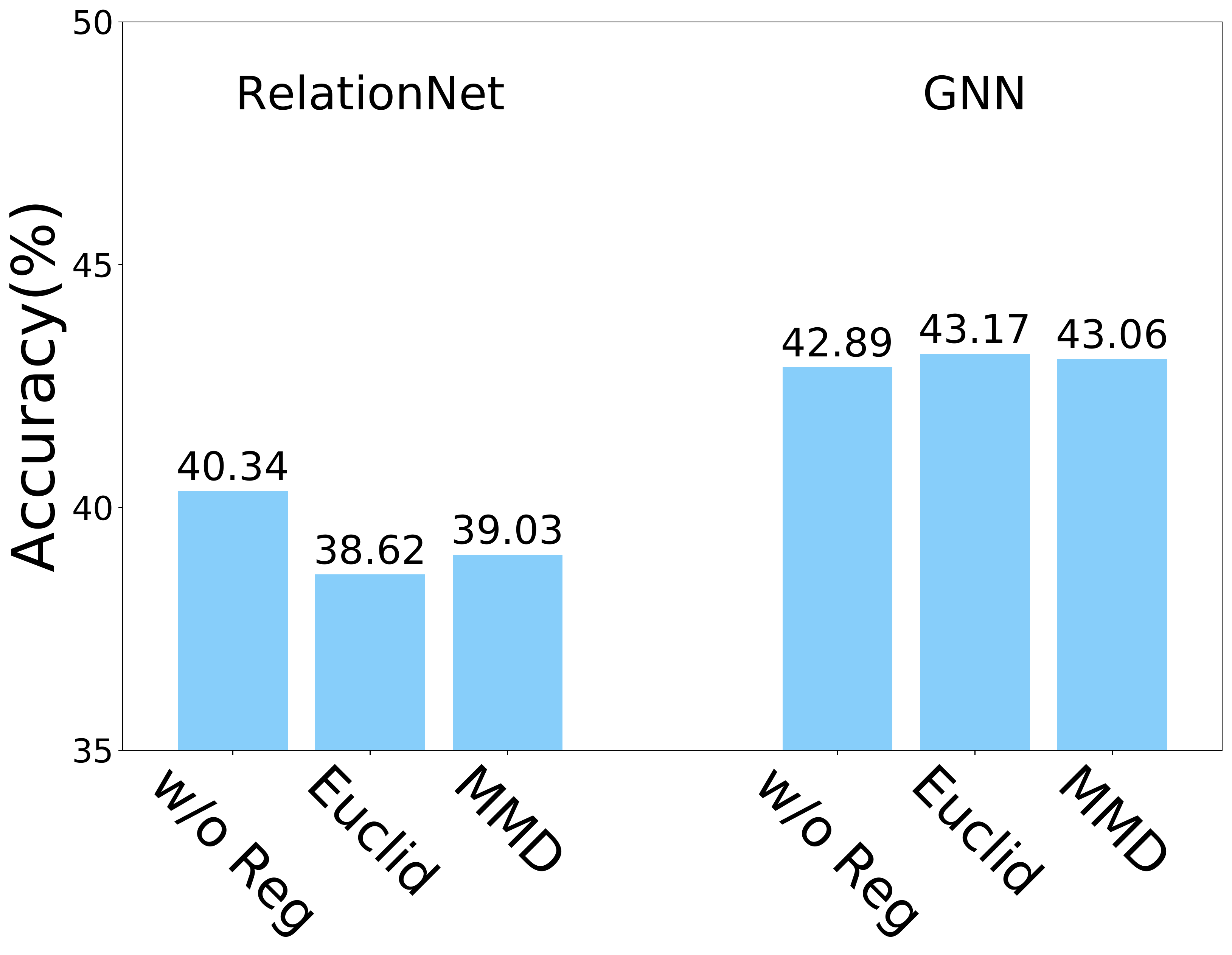}
        \label{fig3a}}
    \subfigure[5-way 5-shot setting]{
        \includegraphics[width=0.225\textwidth]{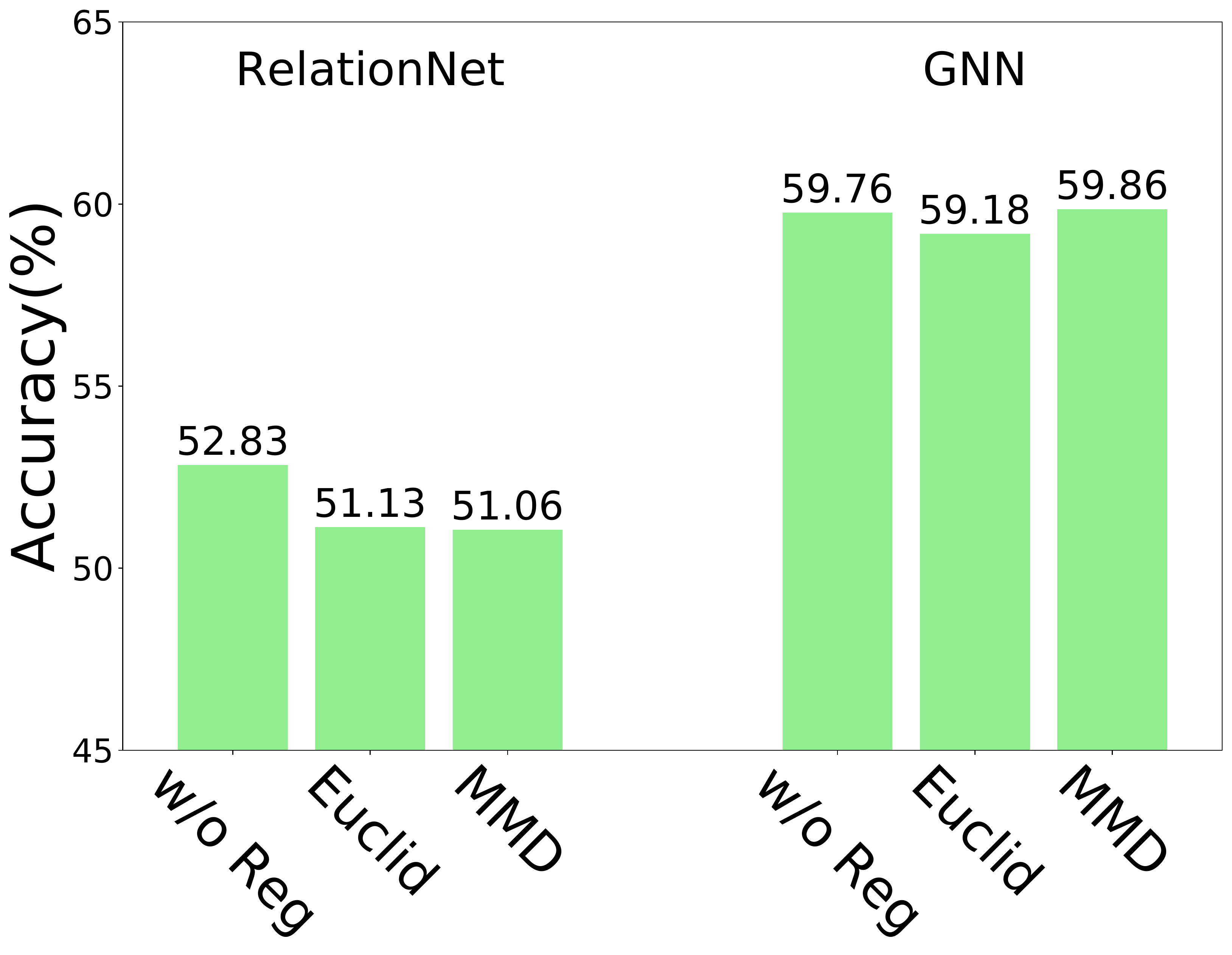}
        \label{fig3b}}
    \caption{Average classification accuracy on eight unseen domains (CUB, Cars, Places, Plantae, CropDiseases, EuroSAT, ISIC and ChestX) under 5-way 1-shot/5-shot setting. It respectively shows the results of the iteration goal without the regularization term ('w/o Reg'), with the sample-wise Euclidean distance regularization term ('Euclid') and with the maximum mean discrepancy (MMD) distance regularization term ('MMD').}
    \label{fig3}
\end{figure}

\begin{table*}[ht]
\centering
\begin{tabular}{ccccccccc}
\toprule
\multicolumn{1}{c}{\multirow{2}{*}{Model}}&\multicolumn{2}{c}{CUB}&\multicolumn{2}{c}{Cars}&\multicolumn{2}{c}{Places}&\multicolumn{2}{c}{Plantae}
\\ \cmidrule(lr){2-3} \cmidrule(lr){4-5} \cmidrule(lr){6-7} \cmidrule(lr){8-9}
\multicolumn{1}{c}{} & \multicolumn{1}{c}{1-shot}&\multicolumn{1}{c}{5-shot} & \multicolumn{1}{c}{1-shot}&\multicolumn{1}{c}{5-shot} & \multicolumn{1}{c}{1-shot}&\multicolumn{1}{c}{5-shot} & \multicolumn{1}{c}{1-shot}&\multicolumn{1}{c}{5-shot} \\
\hline
Fine-tuning              & $43.53_{\pm0.4}$ & $63.76_{\pm0.4}$ & $35.12_{\pm0.4}$ & $51.21_{\pm0.4}$ & $50.57_{\pm0.4}$ & $70.68_{\pm0.4}$ & $38.77_{\pm0.4}$ & $56.45_{\pm0.4}$ \\
RelationNet+ATA$^{\dag}$ & $44.88_{\pm0.4}$ & $66.18_{\pm0.4}$ & $36.44_{\pm0.4}$ & $52.05_{\pm0.4}$ & $52.88_{\pm0.5}$ & $71.40_{\pm0.4}$ & $36.76_{\pm0.4}$ & $54.46_{\pm0.4}$ \\
GNN+ATA$^{\dag}$         & $46.23_{\pm0.5}$ & $\textbf{69.83}_{\pm\textbf{0.5}}$ & $37.15_{\pm0.4}$ & $54.28_{\pm0.5}$ & $54.18_{\pm0.5}$ & $\textbf{76.64}_{\pm\textbf{0.4}}$ & $37.38_{\pm0.4}$ & $58.08_{\pm0.4}$ \\
TPN+ATA$^{\dag}$         & $\textbf{51.89}_{\pm\textbf{0.5}}$ & $\textbf{70.14}_{\pm\textbf{0.4}}$ & $\textbf{38.07}_{\pm\textbf{0.4}}$ & $\textbf{55.23}_{\pm\textbf{0.4}}$ & $\textbf{57.26}_{\pm\textbf{0.5}}$ & $73.87_{\pm0.4}$ & $\textbf{40.75}_{\pm\textbf{0.4}}$ & $\textbf{59.02}_{\pm\textbf{0.4}}$ \\
\midrule
\multicolumn{1}{c}{\multirow{2}{*}{}}&\multicolumn{2}{c}{CropDiseases}&\multicolumn{2}{c}{EuroSAT}&\multicolumn{2}{c}{ISIC}&\multicolumn{2}{c}{ChestX}
\\ \cmidrule(lr){2-3} \cmidrule(lr){4-5} \cmidrule(lr){6-7} \cmidrule(lr){8-9}
\multicolumn{1}{c}{} & \multicolumn{1}{c}{1-shot}&\multicolumn{1}{c}{5-shot} & \multicolumn{1}{c}{1-shot}&\multicolumn{1}{c}{5-shot} & \multicolumn{1}{c}{1-shot}&\multicolumn{1}{c}{5-shot} & \multicolumn{1}{c}{1-shot}&\multicolumn{1}{c}{5-shot} \\
\hline
Fine-tuning          & $73.43_{\pm0.5}$ & $89.84_{\pm0.3}$ & $66.17_{\pm0.5}$ & $81.59_{\pm0.3}$ & $34.60_{\pm0.3}$ & $\textbf{49.51}_{\pm\textbf{0.3}}$ & $\textbf{22.13}_{\pm\textbf{0.2}}$ & $\textbf{25.37}_{\pm\textbf{0.2}}$ \\
RelationNet+ATA$^{\dag}$ & $74.61_{\pm0.5}$ & $90.80_{\pm0.3}$ & $66.18_{\pm0.5}$ & $81.92_{\pm0.3}$ & $32.96_{\pm0.3}$ & $46.99_{\pm0.3}$ & $\textbf{22.24}_{\pm\textbf{0.2}}$ & $\textbf{25.69}_{\pm\textbf{0.2}}$ \\
GNN+ATA$^{\dag}$         & $75.41_{\pm0.5}$ & $\textbf{95.44}_{\pm\textbf{0.2}}$ & $68.62_{\pm0.5}$ & $\textbf{89.64}_{\pm\textbf{0.3}}$ & $\textbf{34.94}_{\pm\textbf{0.4}}$ & $\textbf{49.79}_{\pm\textbf{0.4}}$ & $\textbf{22.15}_{\pm\textbf{0.2}}$ & $25.08_{\pm0.2}$ \\
TPN+ATA$^{\dag}$         & $\textbf{82.47}_{\pm\textbf{0.5}}$ & $93.56_{\pm0.2}$ & $\textbf{70.84}_{\pm\textbf{0.5}}$ & $85.47_{\pm0.3}$ & $\textbf{35.55}_{\pm\textbf{0.4}}$ & $\textbf{49.83}_{\pm\textbf{0.3}}$ & $\textbf{22.45}_{\pm\textbf{0.2}}$ & $24.74_{\pm0.2}$ \\
\bottomrule
\end{tabular}
\caption{Few-shot classification accuracy($\%$) of 5-way 1-shot/5-shot tasks trained with the mini-ImageNet dataset and fine-tuned with the augmented support dataset from the unseen tasks. $\dag$ stands for using the fine-tuning method described in the Section \ref{fine_tune}.}
\label{fine}
\end{table*}

\subsection{Evaluation for Adversarial Task Augmentation}
We apply the adversarial task augmentation module to the RelationNet, the GNN, and the TPN models to evaluate its effect on improving the cross-domain generalization ability of the meta-learning models, and compare it with \cite{DBLP:conf/iclr/TsengLH020} which adds the feature-wise transformation layers to the feature extractor and \cite{sun2020explanation} which uses explanation-guided training. All models are trained and tested in the same environment for the fair comparison and the results are shown in Table \ref{SOTA}.

We can observe that with our adversarial task augmentation module, the cross-domain few-shot classification accuracy of the meta-learning models is consistently and significantly improved. And compared with \cite{DBLP:conf/iclr/TsengLH020} and \cite{sun2020explanation}, our method achieves comparable or more significant improvement, which means that adaptively enhancement of different inductive bias is more effective than enhancing artificially determined inductive bias. Moreover, applying the feature-wise transformation layers even harms the cross-domain generalization performance of the TPN model, while our method is still effective, which means that our method is more general, not just suitable for the metric-based meta-learning models (the RelationNet and the GNN models).

\subsection{Ablation Study}\label{AS}
\paragraph{Effect of the random convolution.}As aforementioned, we use the random convolution for auxiliary task augmentation. Here we study the effect it brings. Figure \ref{fig2} shows the average few-shot classification accuracy on eight unseen domains without random convolution and that obtained by complete method. As we can see, without the random convolution, our method still improves the cross-domain generalization ability of the meta-learning models, and outperforms '+FT' \cite{DBLP:conf/iclr/TsengLH020} and '+LRP' \cite{sun2020explanation}. Using the random convolution can achieve further improvements.

\paragraph{Is the regularization term useful?}In the Section \ref{sec32}, we remove the regularization term $-\gamma d(T,T_0)$ from the iteration goal and here we will show it is reasonable. We consider two common candidates for distance $d(T,T_0)$ and find that they do not bring benefits. As we assumed, the label composition of few-shot classification tasks is the same as each other, so the distance between task $T$ and $T_0$ depends on the samples $X$ and $X_0$. Let the feature vectors of $X$ and $X_0$ are $F=\{f^i\}_{i=1}^N$ and $F_0=\{f_0^i\}_{i=1}^N$ with $N=C\times K+Q$. The first candidate is the direct sample-wise Euclidean distance, i.e., $d(T,T_0)=\frac{1}{N}\sum_{i=1}^N\|f^i-f_0^i\|_2^2$ and the second candidate is the maximum mean discrepancy (MMD) distance, i.e., $d(T,T_0)=\|\frac{1}{N}\sum_{i=1}^Nf^i-\frac{1}{N}\sum_{i=1}^Nf_0^i\|_2^2$. Figure \ref{fig3} shows the average few-shot classification accuracy on eight unseen domains without the regularization term, with the sample-wise Euclidean distance regularization term and with the maximum mean discrepancy (MMD) distance regularization term. We set the hyper-parameter $\gamma=1$ and do not use the random convolution for the clear comparison. As we can see, using the regularization term does not bring obvious benefits or even is harmful, which shows that early stopping has already imposed enough constraints, and using the regularization term leads to excessive limits.

\subsection{Comparison with Fine-tuning}\label{fine_tune}
\cite{guo2020broader} shows that in the cross-domain few-shot classification problem, traditional pre-training and fine-tuning outperform the meta-learning models. Here we re-examine this phenomenon through a different fair comparison, i.e., using data augmentation while solving an unseen task. Given an unseen task $T$ consisting of $C\times K$ support samples and $Q$ query samples, for the fine-tuning, we use the pre-trained feature extractor as initialization and a fully connected layer as the classification head. For each epoch, we generate $15$ pseudo samples for each class based on the support samples using the data augmentation method from \cite{yeh2020large} and use these $C\times15$ pseudo samples and the support samples for fine-tuning where we use the SGD optimizer with the learning rate 0.01 and the momentum 0.9 as in \cite{guo2020broader}. For the meta-learning models, we use the parameters trained on the mini-ImageNet with our adversarial task augmentation method as the initialization and adapt the meta-learning models to the same $C\times(K+15)$ samples as above at each iteration where the $C\times 15$ pseudo samples are used as the pseudo query set, and we use the Adam optimizer with the learning rate 0.001. All models are fine-tuned for 30 (or 50) epochs in the 5-way 1-shot (or 5-shot) tasks. Since all models use the same amount of target domain data when solving each unseen task, it is a fair comparison. The results are shown in Table \ref{fine}. As we can see, the meta-learning models with our adversarial task augmentation module significantly outperform the traditional pre-training and fine-tuning even under domain shift.

\section{Conclusion}
In this paper, we aim to design a new method that can improve the cross-domain generalization capability of meta-learning models in the cross-domain few-shot learning. For this, we consider the worst-case problem around the source task distribution $\mathcal{T}_0$, and propose a plug-and-play inductive bias-adaptive task augmentation method, which significantly improves the cross-domain few-shot classification capability of various meta-learning models, and outperforms the existing works. This is the first work to achieve the above objective by generating ‘challenging’ virtual tasks. We also compare the meta-learning models with pre-training and fine-tuning under the same settings, and find that the meta-learning models with our method outperform the fine-tuning under domain shift.

\bibliographystyle{named}
\bibliography{ijcai21}
\end{document}